\documentclass[supp]{new_tlp}
\usepackage{supp}

\title[Syntax Analysis and Query Evaluation]{Application of Methods for Syntax Analysis of Context-Free Languages to Query Evaluation of Logic Programs}
\author[Heike Stephan]{Heike Stephan\\
	Martin-Luther-Universit\"at Halle-Wittenberg, 
	Institut f\"ur Informatik, 
	06099 Halle (Saale), Germany\\
	heike.stephan@informatik.uni-halle.de}
\pubauthor{Stephan}
\submitted{\today}
\revised{\today}
\accepted{\today}

\begin{document}
\maketitle

\begin{abstract}
My research goal is to employ a parser generation algorithm based on the Earley parsing algorithm to the evaluation and compilation of queries to logic programs, especially to deductive databases. By means of partial deduction, from a query to a logic program a parameterized automaton is to be generated that models the evaluation of this query. This automaton can be compiled to executable code; thus we expect a speedup in runtime of query evaluation.\\

An extended abstract/ full version of a paper accepted to be presented at the Doctoral Consortium of the 30th International Conference on Logic Programming (ICLP 2014), July 19-22, Vienna, Austria
\end{abstract}


\section{Introduction and problem description}
Deductive databases and related rule-based systems are of increasing interest nowadays (ontologies/ semantic web, artificial intelligence, business rules). Especially when large amounts of data have to be processed, an optimal runtime performance is crucial. This can be achieved by compiling the intensional database beforehand, ideally combined with partial evaluation of queries to the deductive database, also known as partial deduction. In this area, lots of work has already been done (see e.\,g.\ \cite{partDeductKomorowski}). 
Runtime performance can also be improved by tabling methods, also known as memoing or memoization techniques (e.\,g.\ the OLDT method \cite{oldtTamakiSato}), which reuse answers to equivalent subgoals and avoid their recomputation. 

One approach that has not yet been taken full advantage of in the past is to make use of the structural similarity of sets of horn clauses, which form classic logic programs, and context-free grammars (see \cite{ullman} for a good description of this connection). Especially for deductive databases the relationship is quite obvious: Intensional predicates correspond to nonterminal symbols and extensional predicates to terminal symbols of a grammar. A query to the database marks a start point corresponding to the start symbol of a grammar. With this, a query to a logic program can be seen as a call to a context-free grammar to produce all words of the language. A parser can be easily modified to fulfil this task instead of consuming words.

The idea is to modify existing, powerful parsing and parser generation algorithms so that they can process logic programs. The rich knowledge of parser generation is thus of great value for the task of compiling queries to logic programs: A parser for a deterministic context-free language can move through a word 
and decide on its acceptance efficiently because 
all grammar derivations that are applicable when a new terminal symbol is read are compiled into the parser; a similar performance for a compiled query to a logic program is desirable. The existing query-evaluating or query-compiling methods look only at single or few similar rules; with applying parser-generation methods, all rules derivable using a database fact can be processed at once. For this reason, we can expect to be better than existing query-evaluation/ query-compilation methods if we employ parsing algorithms for query evaluation.

\section{Background and overview of the existing literature}
Adapting parsing algorithms to sets of horn clauses has especially been done for definite clause grammars (DCGs)---and so remained in the community of computational linguistics---but the ideas may principally be transferred to logic programs in general. 
One option is to use algorithms related to the LR(k) algorithm developed by {\sc Knuth} \cite{knuth}. The LR(k) algorithm is well known and widely used in compiler construction (for a detailed description, see a textbook on compiler construction as e.\,g.\ \cite{aho}). An adaptation to DCGs is presented e.\,g.\ by {\sc Nilsson} in \cite{aid}. Here, a logic program is first reduced to its underlying con\-text\--free grammar for which a LR(k) parser can be generated in traditional manner. The predicate arguments are added during parsing via an argument stack. Algorithms related to the LR(k) algorithm are attractive because, in contrast to pure top-down algorithms, they can be used on left recursive grammars. The main drawback is that the LR(k) algorithm heavily relies on looking ahead one or more characters of the input string in order to guarantee determinism of the generated parser. For the execution of logic programs no input string exists so that the class of programs for which a deterministic execution model can be created is quite small. Of course, one can choose to accept nondeterminism and still profit by being able to cope with left recursion.

Another useful parsing algorithm for evaluating a logic program is Earley Deduction by {\sc Pereira} and {\sc Warren} \cite{pereiraWarren}. This method is inspired by the Earley Algorithm \cite{earley}, a LR parsing algorithm derived from {\sc Knuth}'s LR(k) parser generation algorithm and suitable for all con\-text\--free grammars. 
In contrast to the LR(k) algorithm, Earley Deduction is completely independent of looking ahead input characters. This makes Earley Deduction more attractive for the use in query evaluation than methods based on the LR(k) algorithm.

{\sc Porter} \cite{porter,porter2} tested Earley Deduction in the context of deductive databases and devised several optimizations for the derivation process. He introduced the notion of the \emph{schema} of a rule which makes rules comparable independent of their actual data values. By this means, rules with the same schema can be collected in one data structure and indexed for easy access. 
Additionally, he suggested to precompile derivation steps 
to a set of simple instructions which can be applied to sets of rules. The compilation is done at execution time and depends on the actual data values of the rules with which the comparison or reduction has to be performed. Nevertheless, {\sc Porter} found that his optimizations significantly increase execution speed compared to na\"ive Earley Deduction. 

\section{Goal of the research}
The goal of research is to construct an automaton that models the evaluation process for a logic program in a similar way as a parser is generated automatically from a given grammar. 
The Ear\-ley\--based methods are less restricted than the LR(k) related algorithms for parser generation and will therefore serve as a base for the method to be developed. The following problems have to be solved on the way to this goal:
\begin{enumerate}
	\item In contrast to {\sc Knuth}'s LR(k) algorithm, neither the Earley parsing algorithm nor Earley Deduction are parser generation algorithms. There is no description of algorithms that generate parsers that execute the Earley parsing algorithm or Earley Deduction. 
	\item In order to have a chance to perform better than existing tabling methods, we have to precompile as much of the query evaluation process as possible. This includes especially those derivations that are possible without retrieving new data from the extensional database. In {\sc Knuth}'s LR(k) algorithm and in the Earley parsing algorithm, but not in Earley Deduction, the parser uses states that correspond to sets of partially processed grammar productions derivable after reading one terminal from the input string; reading a terminal leads to a state transition. This behaviour is also desirable for Earley Deduction. 
	\item A parser or parser generator for a DCG---and thus also a query evaluator generator for a logic program---has to cope with the problem of predicate arguments, which are not present in context-free grammars. Not only the status of the evaluation process has to be stored in a state, but also a part of the data from the extensional database. This leads to the unusual concepts of a parameterized state and a parameterized automaton, where the parameters are placeholders for data values retrieved at runtime.
\end{enumerate}

By solving these problems, the state\--tran\-si\-tion feature of the original Earley Algorithm will be combined with the derivation of rules as in Earley Deduction 
to construct a generator for an automaton modeling the query evaluation process. So classical concepts of parser generation are employed for obtaining efficient means to compile a logic program into executable code, which implements the generated automaton. 

The states of the automaton correspond to sets of rules derivable after reading one fact from the extensional database; input of an extensional database fact leads to a state transition. The idea is to represent such a rule set in a way that maps those parts of the rules that depend only on the program to constant symbols, leaving the data values as variables or parameters that are bound at runtime. 
The derivation operations are basically the same for different facts of the same extensional relation and can be computed at compile time. Thus several derivation steps can be processed at once and, additionally, reduced to a sequence of variable assignments and to few comparisons. 

With this we expect to be able to speed up execution by transferring as many actions as possible to the compilation phase. The execution of a program compiled in this way is expected to be faster than execution with Prolog, the Magic Set method \cite{magicSets} or tabling methods, at least in the general case. 

\section{Preliminary results accomplished}
First results have already been published together with my advisor, Prof. S. Brass in \cite{wlp12} and \cite{brass2013variant}. The Earley Deduction method is enhanced by a state transitions and by a generator for a parameterized finite automaton. It can be applied to a non-recursive or left\--re\-cur\-sive Datalog program. 
The execution of the generated automaton performs query evaluation on the given extensional database. The states of this automaton can also be viewed as new Datalog predicates and the state transitions as Datalog rules, so we have a rewritten Datalog program. A bottom-up evaluation of this generated program corresponds to the execution of the parameterized automaton.

\section{Current status of the research}

Currently I am preparing for publication a new version of the method that is able to deal with Datalog programs of the kind of the Same-Generations Program. I also have almost finished thoughts on a version of the method that is applicable to general logic programs, including negation as finite failure/stratified negation. 

\section{Open issues and expected achievements}

There are several ideas on how the versions of the Earley-based compilation methods can be implemented; there will have to be comparisons between those implementations as well as to other established implementations of Datalog and Prolog, especially those using tabling and Magic Sets. The expected improvement of runtime performance must be demonstrated. Furthermore, I am particularly interested in reapplying my compilation method to parser generation and examine how the logic programs that can be processed with the current version are connected to grammars for deterministic context-free languages. If those connections are sufficient, I intend to use the method as parser generator for attributed grammars.

\bibliographystyle{acmtrans}
\bibliography{literatur.bib}
\end{document}